\documentclass[journal]{IEEEtran}
\usepackage{amsmath,graphicx}
\usepackage{amssymb}
\usepackage[printonlyused,nolist]{acronym}
\usepackage{pgfplots}
\usepackage{cite}
\usepackage{pgfplotstable}
\usepackage[utf8]{inputenc}
\usepackage{tikz}
\usepackage{psfrag}
\usepackage{textcomp}
\usepackage{flushend} 
\usepackage{colortbl}

\tikzset{dashdot/.style={dash pattern=on .8pt off 3pt on 4pt off 3pt}}

  \usepackage{subfigure} 

\pgfplotsset{compat=1.3}

\begin{acronym}
\acro{NLMS}{normalized least mean square}
\acro{AEC}{acoustic echo cancellation}
\acro{FIR}{finite impulse response}
\acro{RIR}{room impulse response}
\acro{EM}{expectation-maximization}
\acro{MMSE}{minimum mean square error}
\acro{PDF}{probability density function}
\acro{KLMS}{Kalman-based LMS algorithm}
\end{acronym}

\begin{document}
\title{The NLMS~algorithm~with~time-variant~optimum\\stepsize~derived~from~a~Bayesian~network~perspective}
\author{Christian~Huemmer,~\IEEEmembership{Student Member,~IEEE},
	Roland~Maas,
        Walter~Kellermann,~\IEEEmembership{Fellow,~IEEE}
}

\markboth{IEEE Signal Processing Letters,~Vol.~XX, No.~X, Nov.~2014}%
{Huemmer \MakeLowercase{\textit{et al.}}: A new derivation of the time-varying NLMS stepsize parameter from a Bayesian network perspective}



\maketitle

\begin{abstract}
In this article, we derive a new stepsize adaptation for the normalized least mean square algorithm (NLMS) by describing the task of linear acoustic echo cancellation from a Bayesian network perspective.
Similar to the well-known Kalman filter equations, we model the acoustic wave propagation from the loudspeaker to the microphone by a latent state vector and define a linear
observation equation (to model the relation between the state vector and the observation) as well as a linear process equation (to model the temporal progress of the state vector).
Based on additional assumptions on the statistics of the random variables in observation and process equation, we apply the expectation-maximization (EM) algorithm to
derive an \mbox{NLMS-like} filter adaptation. By exploiting the conditional independence rules for Bayesian networks, we reveal that the resulting EM-NLMS algorithm has a stepsize update equivalent to the optimal-stepsize calculation proposed by Yamamoto and Kitayama
in 1982, which has been adopted in many textbooks.
As main difference, the instantaneous stepsize value is estimated in the M~step of the EM algorithm (instead of being approximated by artificially extending the acoustic echo path).
The EM-NLMS algorithm is experimentally verified
for synthesized scenarios with both, white noise and male speech as input signal.
\end{abstract}

\IEEEpeerreviewmaketitle

\section{Introduction}
\label{sec:Intro}
\begin{figure}[b^]
\centering
\psfrag{A}[c][c] {$\mathbf{x}_n$}
\psfrag{B}[c][c] {$\mathbf{h}_n$}
\psfrag{C}[c][c] {$v_n$}
\psfrag{D}[c][c] {$d_n$}
\psfrag{F}[c][c] {$\hat{d}_n$}
\psfrag{G}[c][c] {$e_n$}
\psfrag{E}[c][c] {$\mathbf{\hat{h}}_{n-1}$}
\includegraphics[width=0.3\textwidth]{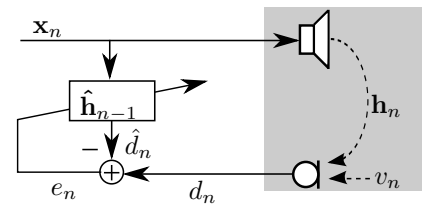}
\caption{System model for linear \ac{AEC} with RIR vector $\mathbf{h}_n$}
\label{fig:NONlinAEC}
\end{figure}

\IEEEPARstart{M}{achine} learning techniques have been widely applied to signal processing tasks since decades \cite{frey2005,adali2011}. 
For example, directed graphical models, termed Bayesian networks, have shown to provide a powerful framework for modeling causal probabilistic relationships between random variables~\cite{bilmes2005,ITG6,wainwright2007,barber2010,ChinaSIP7}. In previous work,
the update equations of the Kalman filter and the \ac{NLMS} algorithm have already been
derived from a Bayesian network perspective based on a linear relation between the latent \ac{RIR} vector and the observation~\cite{maas2014,Bishop}.\\
The \ac{NLMS} algorithm is one of the most-widely used adaptive algorithms in speech signal processing and a variety of stepsize adaptation 
schemes has been proposed to
improve its system identification performance~\cite{Variable1982,Variable1992,Variable1997,Variable2000,shin2004,Variable2006,KLMS,Variable2008,paleologu2009,hwang2009,Variable2012,zhao2013}.
In this article, we derive a novel \ac{NLMS}-like filter adaptation (termed EM-NLMS algorithm)
by applying the \ac{EM} algorithm to a probabilistic model for linear system identification.
Based on the conditional independence rules for Bayesian networks, it is shown that the normalized stepsize of the EM-NLMS algorithm
is equivalent to the one proposed in~\cite{Variable1982}, which is now commonly accepted as optimum NLMS stepsize rule, see e.g. \cite{AdaptiveFilter}.
As the main difference relative to~\cite{Variable1982} , the normalized stepsize~is~here estimated as part of the \ac{EM} algorithm instead of being approximated by artificially extending the acoustic echo~path.
For a valid comparison, we review the algorithm~of~\cite{Variable1982}~for~the linear \ac{AEC} scenario shown in~Fig.~\ref{fig:NONlinAEC}.
The acoustic path between loudspeaker and microphone at time $n$ is modeled by the linear~\ac{FIR}~filter
\begin{equation}
\mathbf{h}_n=[h_{0,n},h_{1,n},...,h_{M-1,n}]^T
\label{equ:hGen}
\end{equation}
with time-variant coefficients $h_{\kappa,n}$, where ${\kappa=0,...,M-1}$.
The observation equation models the microphone sample $d_n$:
\begin{equation}
d_n =   \mathbf{x}^T_n\mathbf{h}_n  + v_n,
\label{equ:ObservSpe}
\end{equation}
with the additive variable $v_n$ modeling near-end interferences and the observed
input signal vector \mbox{$\mathbf{x}_n=[x_n,x_{n-1},...,x_{n-M+1}]^T$} capturing the time-domain samples $x_n$.
The iterative estimation of the \ac{RIR} vector by the adaptive \ac{FIR} filter $\mathbf{\hat{h}}_{n}$ is realized by the update rule
\begin{equation}
\mathbf{\hat{h}}_{n} = \mathbf{\hat{h}}_{n-1} + \lambda_n \mathbf{x}_n  e_n,
\label{equ:UpdNLMS}
\end{equation}
with the stepsize $\lambda_n$ and the error signal
\begin{equation} 
e_n = d_n-\mathbf{x}_n^T\mathbf{\hat{h}}_{n-1}
\label{equ:Error}
\end{equation}
relating the observation $d_n$ and its estimate $\hat{d}_n=\mathbf{x}_n^T\mathbf{\hat{h}}_{n-1}$. 
In~\cite{Variable1982}, the optimal choice of $\lambda_n$ has been approximated as:
\begin{equation}
\lambda_{n} \approx \frac{1}{M} \frac{\mathcal{E}\{ ||\mathbf{h}_n - \mathbf{\hat{h}}_{n-1}  ||^2_2 \}}{\mathcal{E}\{e_n^2\}},
\label{equ=alphaSchul}
\end{equation}
where $||\cdot||_2$ denotes the Euclidean norm and $\mathcal{E}\{\cdot\}$ the expectation operator.
As the true echo path $\mathbf{h}_n$ is unobservable, so that the numerator in (\ref{equ=alphaSchul}) cannot be computed, $\lambda_n$ is further approximated by introducing a delay of
$N_T$ coefficients to the echo path $\mathbf{h}_n$. Moreover, a recursive approximation of the denominator in (\ref{equ=alphaSchul}) is applied
using the forgetting factor~$\eta$~\mbox{\cite{AdaptiveFilter,NMLSapprox2}}. The resulting stepsize approximation
\begin{equation}
\lambda_{n} \approx  \frac{1}{ N_T}   \frac{  \sum\limits_{\kappa=0}^{N_T-1} \hat{h}^2_{k,n-1} }{ (1-\eta) e_n^2 + \eta \mathcal{E}\{e_{n-1}^2\} }
\label{equ=alphaSchulApp}
\end{equation}
leads to oscillations which have to be addressed by limiting the absolute value~of~$\lambda_{n}$~\cite{NLMSschultheis}.
\renewcommand{\arraystretch}{2.5}
\begin{table}
\caption{Relation between the NLMS algorithm following~\cite{Variable1982} and the proposed EM-NLMS algorithm}
\centering
  \begin{tabular}[T]{|c|c c|}
  \hline
  & NLMS algorithm \cite{Variable1982} \;&\; EM-NLMS algorithm \\[0.5mm]
\hline
  Norm. stepsize $\lambda_n$ & (\ref{equ=alphaSchul}) \qquad  \quad& \qquad E step \\[1mm]
  \hline
  Estimation of $\lambda_n$ & (\ref{equ=alphaSchulApp})\qquad \quad & \qquad M step\\[1mm]  \arrayrulecolor{black}\hline
  \end{tabular}\\[-17mm]
  \begin{tikzpicture}
      \node at (1,4.1) {$\hspace{25mm}$ \text{\footnotesize equivalent to}};
      \node at (1,3.8) {$\hspace{25mm}$ \text{\footnotesize  (Section~\ref{sec:Comp}})};
      \node at (1,3.22) {$\hspace{25mm}$ \text{\footnotesize replaced by}};
      \node at (1,2.92) {$\hspace{25mm}$ \text{\footnotesize  (Subsec.~\ref{cha:ParticleModif2}})};
\end{tikzpicture}\\[-2mm]
\label{tab:Comp}
\end{table}
In this article, we derive~the \mbox{EM-NLMS} algorithm which applies the filter update of~(\ref{equ:UpdNLMS}) using the stepsize in~(\ref{equ=alphaSchul}),~where $\lambda_n$ is 
estimated in the~M~Step~of the EM algorithm instead of being approximated by using~(\ref{equ=alphaSchulApp}).\newpage
\noindent This article is structured as follows: In Section~\ref{sec:BayesianNetwork}, we propose a probabilistic model for the linear \ac{AEC} scenario of Fig.~\ref{fig:NONlinAEC} and derive the EM-NLMS algorithm, which is revealed
in Section~\ref{sec:Comp} to be similar to the NLMS algorithm proposed~in~\cite{Variable1982}.
As main difference~(cf. Table~\ref{tab:Comp}), the stepsize is estimated in the M Step of the EM algorithm instead of being approximated by artificially extending the acoustic echo path.
In Section~\ref{sec:exper}, the EM-NLMS algorithm is experimentally verified for synthesized scenarios
with both, white noise and male speech as input signal.
Finally, conclusions are drawn in Section~\ref{sec:conclu}.
\section{The EM-NLMS algorithm for linear AEC}
\label{sec:BayesianNetwork}
Throughout this article, the Gaussian \ac{PDF} of a real-valued length-$M$ vector $\mathbf{z}_n$ 
with mean vector~$\boldsymbol{\mu}_{\mathbf{z},n}$ and covariance matrix $\mathbf{C}_{\mathbf{z},n}$ is denoted~as
\begin{equation} 
\begin{split}
&\mathbf{z}_n\sim  \mathcal{N} \{ \mathbf{z}_n | \boldsymbol{\mu}_{\mathbf{z},n},\mathbf{C}_{\mathbf{z} ,n} \} \\
=& \frac{ |\mathbf{C}_{\mathbf{z} ,n}|^{-1/2}}{(2 \pi)^{M/2}} \exp \left\{ -\frac{ (\mathbf{z}_n-\boldsymbol{\mu}_{\mathbf{z},n})^T \mathbf{C}_{\mathbf{z} ,n}^{-1}(\mathbf{z}_n-\boldsymbol{\mu}_{\mathbf{z},n})}{2} \right\},
\end{split}
\label{equ:GaussianPDF}
\end{equation} 
where $|\cdot|$ represents the determinant of a matrix.
Furthermore,
\mbox{$\mathbf{C}_{\mathbf{z},n}=C_{\mathbf{z},n} \mathbf{I}$} (with identity matrix~$\mathbf{I}$) implies 
the elements of $\mathbf{z}_n$ to be mutually statistically independent and of equal variance $C_{\mathbf{z},n}$.
\subsection{Probabilistic AEC model}
\begin{figure}[!t]
\centering
\psfrag{A1}[c][c]{$\mathbf{w}_1$} \psfrag{A2}[c][c]{$\mathbf{w}_2$} \psfrag{An1}[c][c]{$\mathbf{w}_{n-1}$} \psfrag{An}[c][c]{$\mathbf{w}_n$} 
\psfrag{b1}[c][c]{$\mathbf{h}_1$} \psfrag{b2}[c][c]{$\mathbf{h}_2$} \psfrag{bn1}[c][c]{$\mathbf{h}_{n-1}$} \psfrag{bn}[c][c]{$\mathbf{h}_n$}
\psfrag{c1}[c][c]{$d_1$} \psfrag{c2}[c][c]{$d_2$} \psfrag{cn1}[c][c]{$d_{n-1}$} \psfrag{cn}[c][c]{$d_n$}
\psfrag{d1}[c][c]{$v_1$} \psfrag{d2}[c][c]{$v_2$} \psfrag{dn1}[c][c]{$v_{n-1}$} \psfrag{dn}[c][c]{$v_n$}
\includegraphics[width=0.34\textwidth]{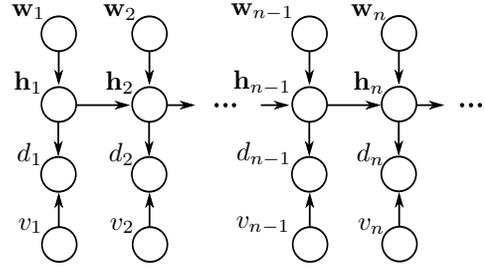}
\caption{Bayesian network for linear AEC with latent state vector $\mathbf{h}_n$}
\label{fig:ParticleModel}
\end{figure}

To describe the linear \ac{AEC} scenario of Fig.~\ref{fig:NONlinAEC} from a Bayesian network perspective, we model
the acoustic echo path as a latent state vector $\mathbf{h}_n$ identically defined as in~(\ref{equ:hGen})
and capture uncertainties (e.g. due to the limitation to a linear system with a finite set of coefficients) by the additive uncertainty~$\mathbf{w}_n$. Consequently, the linear process equation and the linear observation equation,
\begin{equation}
\mathbf{h}_n= \mathbf{h}_{n-1} + \mathbf{w}_n \quad \text{and} \quad d_n =   \mathbf{x}^T_n\mathbf{h}_n  + v_n,
\label{equ:TranEq}
\end{equation}
can be jointly represented by the graphical model shown in Fig.~\ref{fig:ParticleModel}.
The directed links express statistical dependencies between the nodes and random variables, such as $v_n$, are marked as circles.
We make the following assumptions on the \ac{PDF}s of the random variables~in~Fig.~\ref{fig:ParticleModel}:
\begin{itemize}
 \item 
The uncertainty $\mathbf{w}_n$ is normally distributed with mean vector $\mathbf{0}$ (of zero-valued entries) and variance~$C_{\mathbf{w},n}$:
\begin{equation}
\mathbf{w}_n \sim \mathcal{N} \{ \mathbf{w}_n | \mathbf{0}, \mathbf{C}_{\mathbf{w} ,n}\} ,\quad \mathbf{C}_{\mathbf{w} ,n} = C_{\mathbf{w},n} \mathbf{I}.
\label{equ:ChannelUncert}
\end{equation}
\item
The microphone signal uncertainty $v_n$ is assumed to be normally distributed with variance $C_{v, n}$ and zero mean:
\begin{equation}
v_n \sim \mathcal{N} \{ v_n | 0, C_{v, n}\}.
\label{equ:micUnc}
\end{equation}
\item 
The posterior distribution $p \left( \mathbf{h}_n | d_{1:n}\right)$ is defined with mean vector $\boldsymbol{\mu}_{\mathbf{h},n}$, variance~$ C_{\mathbf{h},n}$~and~${d_{1:n}=d_1,...,d_n}$:
\begin{equation}
p \left( \mathbf{h}_n | d_{1:n}\right) = \mathcal{N} \{ \mathbf{h}_n | \boldsymbol{\mu}_{\mathbf{h},n},\mathbf{C}_{\mathbf{h} ,n} \},\quad \mathbf{C}_{\mathbf{h} ,n} = C_{\mathbf{h},n} \mathbf{I}.
\label{equ:Posterior}
\end{equation}
\end{itemize}
\noindent Based on this probabilistic AEC model, we apply the \ac{EM} algorithm consisting of two parts:
In the E~Step, the filter
update is derived based on \ac{MMSE} estimation (Subsection~\ref{cha:ParticleModif}).
In the M~step, we predict the model parameters $C_{v, n+1}$~and~$C_{\mathbf{w}, n+1}$ to estimate the adaptive stepsize value $\lambda_{n+1}$~(Subsection~\ref{cha:ParticleModif2}).
%
\subsection{E step: Inference of the state vector}
\label{cha:ParticleModif}
The \ac{MMSE} estimation of the state vector
identifies the mean vector of the posterior distribution as estimate $\mathbf{\hat{h}}_n$:
\begin{equation}
\mathbf{\hat{h}}_n = \underset{\mathbf{\tilde{h}}_n}{\operatorname{argmin}} \; \mathcal{E} \{ ||\mathbf{\tilde{h}}_n-\mathbf{h}_n ||_2^2 \} = \mathcal{E} \{ \mathbf{h}_n | d_{1:n} \}=\boldsymbol{\mu}_{\mathbf{h},n}.
\label{equ:mes}
\end{equation}
Due to the linear relations between the variables in (\ref{equ:ObservSpe})~and~(\ref{equ:TranEq}),
and under the restrictions to a linear estimator of $\mathbf{\hat{h}}_n$ and normally distributed random variables, the \ac{MMSE} estimation is analytically tractable~\cite{Bishop} .
Exploiting the product rules for linear Gaussian models and conditional independence of the Bayesian network in Fig~\ref{fig:ParticleModel}, the filter update can be derived as a special case of the Kalman filter equations \cite[p. 639]{Bishop}:
\begin{equation} 
\mathbf{\hat{h}}_n = \mathbf{\hat{h}}_{n-1} + \boldsymbol{\Lambda}_n \mathbf{x}_n  e_n,
\label{equ:LDSupdate}
\end{equation}
with the stepsize matrix
\begin{equation} 
\boldsymbol{\Lambda}_n= \frac{ \mathbf{C}_{\mathbf{h} ,n-1} +\mathbf{C}_{\mathbf{w} ,n} }{\mathbf{x}^T_n ( \mathbf{C}_{ \mathbf{h} ,n-1} +\mathbf{C}_{ \mathbf{w} ,n} )\mathbf{x}_n + C_{v, n}}
\end{equation}
and the update of the covariance matrix given as
\begin{equation} 
\mathbf{C}_{\mathbf{h} ,n} = \left( \mathbf{I} -\boldsymbol{\Lambda}_n \mathbf{x}_n \mathbf{x}^T_n \right) (\mathbf{C}_{ \mathbf{h} ,n-1} +\mathbf{C}_{ \mathbf{w} ,n} )
\label{equ:LDSupdateC}.
\end{equation}
By inserting (\ref{equ:ChannelUncert}) and (\ref{equ:Posterior}), we can rewrite the filter update of~(\ref{equ:LDSupdate}) to the filter update defined in (\ref{equ:UpdNLMS}) with the scalar stepsize
\begin{equation}
\lambda_n= \frac{C_{\mathbf{h},n-1} +C_{\mathbf{w},n}}{\mathbf{x}^T_n\mathbf{x}_n (C_{\mathbf{h},n-1} +C_{\mathbf{w},n}) + C_{v,n}}.
\label{equ:Lmabda}
\end{equation}
Finally, the update of $C_{\mathbf{h},n}$ is approximated following (\ref{equ:Posterior}) as
\begin{equation} 
C_{\mathbf{h},n} \stackrel{(\ref{equ:Posterior})}{=} \frac{\text{diag}\{\mathbf{C}_{\mathbf{h},n}\}}{M}\stackrel{(\ref{equ:LDSupdateC})}{=}  \left( 1 -\lambda_n \frac{\mathbf{x}^T_n \mathbf{x}_n}{M} \right) (C_{\mathbf{h},n-1} +C_{\mathbf{w},n}),
\end{equation}
where $\text{diag}\{\cdot\}$ adds up the diagonal elements of a matrix.\\
\noindent Before showing the equality of the stepsize updates in (\ref{equ:Lmabda})~and~(\ref{equ=alphaSchul}) in Section~\ref{sec:Comp}, we propose a new alternative to estimate $\lambda_n$ in (\ref{equ:Lmabda}) by
deriving the updates of the model parameters $C_{\mathbf{w},n}$ and $C_{v,n}$ in the following section.
\subsection{M step: Online learning of the model parameters}
\label{cha:ParticleModif2}
In the M step,  we predict the model parameters for the following time instant.
Although the maximum likelihood estimation is analytically tractable, we apply the \ac{EM}~algorithm to derive an online estimator:
In order to update \mbox{$\theta_{n}=\{ C_{v,n}, C_{\mathbf{w},n}\}$} to  the new parameters \mbox{$\theta^{\text{new}}_{n}=\{ C^{\text{new}}_{v,n}, C^{\text{new}}_{\mathbf{w},n}\}$},
the lower~bound
\begin{equation}
 \mathcal{E}_{\mathbf{h}_{1:n}|\theta_{1:n}} \{ \ln\left( p(d_{1:n}, \mathbf{h}_{1:n}|\theta_{1:n} ) \right)\}  \leq \ln p(d_{1:n}|\theta_{1:n}) ,
\label{equ:bound}
\end{equation}
is maximized, where \mbox{$\theta_{1:n}=\{ C_{v,1:n}, C_{\mathbf{w},1:n}\}$}. For this, the \ac{PDF} $p(d_{1:n}, \mathbf{h}_{1:n}|\theta_{1:n})$ is determined by applying the decomposition rules for Bayesian networks \cite{Bishop}:
\begin{align}
&p(d_{1:n}, \mathbf{h}_{1:n} |\theta_{1:n}) =  p(\mathbf{h}_{n} |\mathbf{h}_{n-1}, C_{\mathbf{w},n} \mathbf{I}) p(d_{n}| \mathbf{h}_{n},C_{v,n})\notag \\
&\cdot \prod_{m=1}^{n-1} p(\mathbf{h}_{m} |\mathbf{h}_{m-1}, C_{\mathbf{w},m} \mathbf{I}) p(d_{m}| \mathbf{h}_{m},C_{v,m}).
\label{equ:JointPDFtot}
\end{align}
Next, we take the natural logarithm ln$(\cdot)$ of $p(d_{1:n}, \mathbf{h}_{1:n} |\theta_{1:n})$, replace $\theta_{n}$ by $\theta^{\text{new}}_{n}$ and maximize the right-hand side of~(\ref{equ:bound}) with respect to~$\theta^{\text{new}}_{n}$: 
\begin{align}
\theta^{\text{new}}_{n} &= \underset{C^{\text{new}}_{\mathbf{w},n}}{\operatorname{argmax}} \; \mathcal{E}_{ \mathbf{h}_{1:n}|\theta_{n}} \{ \ln \left( p(\mathbf{h}_{n} |\mathbf{h}_{n-1}, C^{\text{new}}_{\mathbf{w},n} \mathbf{I})\right)\} \notag \\
&+ \underset{C^{\text{new}}_{v,n}}{\operatorname{argmax}} \; \mathcal{E}_{ \mathbf{h}_{1:n}|\theta_{n}} \{ \ln \left( p(d_{n}| \mathbf{h}_{n},C^{\text{new}}_{v,n})\right)\},
\label{equ:maxDef}
\end{align}
where we apply two separate maximizations starting~with
the estimation of $C^{\text{new}}_{v,n}$ by inserting
\begin{equation}
\ln(p(d_{n}| \mathbf{h}_{n} ,C^{\text{new}}_{v,n})) \stackrel{(\ref{equ:TranEq})}{=} -\frac{\ln(2\pi C^{\text{new}}_{v, n} )}{2} - \frac{(d_n - \mathbf{x}^T_n\mathbf{h}_n)^2}{2 C^{\text{new}}_{v, n}}
\end{equation}
into (\ref{equ:maxDef}). This leads to the instantaneous estimate:
\begin{align}
C^{\text{new}}_{v,n} &= \mathcal{E}_{\mathbf{h}_{1:n}|\theta_{n}} \{ ( d_n - \mathbf{x}^T_n\mathbf{h}_n )^2  \} \\ 
&  = d_n + \mathbf{x}^T_n ( C_{\mathbf{h},n}\mathbf{I}+\mathbf{\hat{h}}_n \mathbf{\hat{h}}^T_n ) \mathbf{x}_n - 2 \mathbf{x}^T_n \mathbf{\hat{h}}_n \\ 
&= ( d_n - \mathbf{x}^T_n\mathbf{\hat{h}}_n )^2 + \mathbf{x}^T_n\mathbf{x}_n C_{\mathbf{h},n}.
\label{equ:UpdateV}
\end{align}
The variance (of the microphone signal uncertainty) $C^{\text{new}}_{v,n}$ in~(\ref{equ:UpdateV}) consists of two components, which can be interpreted as follows~\cite{itg2014}: 
The first term in~(\ref{equ:UpdateV}) is given as the squared error signal after filter adaptation and is influenced by near-end interferences like background noise.
The second term in~(\ref{equ:UpdateV}) depends on the signal energy $\mathbf{x}^T_n\mathbf{x}_n$ and the variance $C_{\mathbf{h},n}$ which implies that it considers uncertainties in the linear echo path model.
Similar to the derivation for $C^{\text{new}}_{v,n}$, we insert  
\begin{flalign}
&\ln( p(\mathbf{h}_{n} |\mathbf{h}_{n-1},C_{\mathbf{w},n} \mathbf{I})) \notag \\[2mm]
&\stackrel{(\ref{equ:TranEq})}{=}-\frac{M\ln(2\pi C^{\text{new}}_{\mathbf{w}, n} )}{2}- \frac{(\mathbf{h}_n-\mathbf{h}_{n-1})^T(\mathbf{h}_n-\mathbf{h}_{n-1})}{2 C^{\text{new}}_{\mathbf{w}, n}}
\end{flalign}
into (\ref{equ:maxDef}),
to derive the instantaneous estimate of $C^{\text{new}}_{\mathbf{w},n}$:
\begin{align}
C^{\text{new}}_{\mathbf{w},n} = \; &  \frac{1}{M} \mathcal{E}_{\mathbf{h}_{1:n}|\theta_{n}} \{ (\mathbf{h}_n-\mathbf{h}_{n-1})^T(\mathbf{h}_n-\mathbf{h}_{n-1})\} \\
\stackrel{(\ref{equ:Posterior})}{=} &C_{\mathbf{h},n} - C_{\mathbf{h},n-1}  + \frac{1}{M} \left( \mathbf{\hat{h}}_n^T\mathbf{\hat{h}}_n - \mathbf{\hat{h}}_{n-1}^T\mathbf{\hat{h}}_{n-1} \right),
\label{equ:UpdateW}
\end{align}
where we employed the statistical independence between $\mathbf{w}_n$ and $\mathbf{h}_{n-1}$.
Equation (\ref{equ:UpdateW}) implies the estimation of $C^{\text{new}}_{\mathbf{w},n}$ as difference of the filter tap autocorrelations between the time instants $n$ and $n-1$.
Finally, the updated values in $\theta^{\text{new}}_{n}$ are used as initialization for the following time step, so that
\begin{equation}
\theta_{n+1} := \theta^{\text{new}}_{n} \; \rightarrow \; C_{\mathbf{w},n+1} := C^{\text{new}}_{\mathbf{w},n}, \;C_{v,n+1} := C^{\text{new}}_{v,n}.
\end{equation}
\section{Comparison between the EM-NLMS algorithm and the NLMS algorithm proposed in \cite{Variable1982}}
\label{sec:Comp}
In this part, we compare the proposed EM-NLMS~algorithm to the NLMS algorithm reviewed in Section~\ref{sec:Intro} and show the equality between the adaptive stepsizes in (\ref{equ=alphaSchul})~and~(\ref{equ:Lmabda}).
We reformulate the stepsize update in (\ref{equ:Lmabda}) by applying the conditional independence rules for Bayesian networks~\cite{Bishop}:
First, we exploit the equalities
\begin{equation} 
\begin{split}
 \mathbf{C}_{\mathbf{h} ,n} &\stackrel{(\ref{equ:Posterior})}{=} C_{\mathbf{h},n} \mathbf{I}\stackrel{(\ref{equ:mes})}{=}\mathcal{E} \{ (\mathbf{h}_n- \mathbf{\hat{h}}_n ) (\mathbf{h}_n- \mathbf{\hat{h}}_n )^T\},\\
  \mathbf{C}_{\mathbf{w} ,n}&\stackrel{(\ref{equ:ChannelUncert})}{=}C_{\mathbf{w},n} \mathbf{I} = \mathcal{E} \{ \mathbf{w}_n \mathbf{w}_n^T\},
\end{split}
\label{equ:varDef}
 \end{equation}
which lead to the following relations:
\begin{align} 
 C_{\mathbf{h},n} &= \frac{\mathcal{E} \{ (\mathbf{h}_n- \mathbf{\hat{h}}_n )^T (\mathbf{h}_n- \mathbf{\hat{h}}_n )\}}{M} = \frac{\mathcal{E}\{||\mathbf{h}_{n} - \mathbf{\hat{h}}_{n}||^2_2\}}{M} ,\notag \\
 C_{\mathbf{w},n} &= \frac{\mathcal{E} \{ \mathbf{w}_n^T \mathbf{w}_n\}}{M} =\frac{ \mathcal{E}\{||\mathbf{w}_{n}||^2_2\}}{M}.
\label{equ:Zwischenschritt}
\end{align}
Second, it can be seen in Fig.~\ref{fig:ParticleModel} that the state vector $\mathbf{h}_{n-1}$ and the uncertainty $\mathbf{w}_n $ are statistically independent as they share a head-to-head relationship with respect to the latent vector $\mathbf{h}_{n}$.
As a consequence, the numerator in (\ref{equ:Lmabda}) can be rewritten~as
\begin{align}
C_{\mathbf{h},n-1} + C_{\mathbf{w},n}    \stackrel{(\ref{equ:Zwischenschritt})}{=} \; & \frac{\mathcal{E}\{||\mathbf{h}_{n-1} - \mathbf{\hat{h}}_{n-1}||^2_2\}}{M} +\frac{ \mathcal{E}\{||\mathbf{w}_{n}||^2_2\}}{M} \notag \\
 \stackrel{(\ref{equ:TranEq})}{=} \; & \frac{\mathcal{E}\{||\mathbf{h}_{n} - \mathbf{\hat{h}}_{n-1}||^2_2\}}{M}.
\label{equ:UmrechnungD}
\end{align}
Finally, we consider the mean of the squared error signal 
\begin{equation}
\mathcal{E}\{e_n^2\}   \stackrel{(\ref{equ:ObservSpe}),(\ref{equ:Error})}{=} \mathcal{E}\{(\mathbf{x}_n^T(\mathbf{h}_n-\mathbf{\hat{h}}_{n-1}) + v_n)^2\},
\label{equ:UmrechnungE1}
\end{equation}
which is not conditioned on the microphone signal $d_n$. By applying the conditional independence rules to the Bayesian network in Fig.~\ref{fig:ParticleModel}, the head-to-head relationship with respect to $d_n$ implies $v_n$ to be statistically independent from 
$\mathbf{h}_{n-1}$ and $\mathbf{w}_{n}$, respectively. Consequently, we can rewrite (\ref{equ:UmrechnungE1}) as:
\begin{align}
\mathcal{E}\{e_n^2\}
  \stackrel{(\ref{equ:micUnc})}{=}  \hspace*{1.2mm} & \hspace*{3mm} \mathbf{x}_n^T \mathcal{E}\{(\mathbf{h}_n-\mathbf{\hat{h}}_{n-1})(\mathbf{h}_n-\mathbf{\hat{h}}_{n-1})^T\}\mathbf{x}_n   + C_{v,n}\notag \\
\stackrel{(\ref{equ:TranEq}),(\ref{equ:varDef})}{=}  \hspace*{-1.2mm} & \hspace*{3mm}\mathbf{x}^T_n\mathbf{x}_n (C_{\mathbf{h},n-1} +C_{\mathbf{w},n}) + C_{v,n}.
\label{equ:UmrechnungE2}
\end{align}
The insertion of (\ref{equ:UmrechnungD})~and~(\ref{equ:UmrechnungE2}) into the stepsize defined in~(\ref{equ:Lmabda}) yields the identical expression for $\lambda_n$ as in (\ref{equ=alphaSchul}).
The main difference of the proposed EM-NLMS algorithm is that the model parameters $C_{\mathbf{h},n}$ and $C_{\mathbf{w},n}$ (and consequently the normalized stepsize $\lambda_n$) are estimated in the M step of the EM algorithm instead of being approximated
using~(\ref{equ=alphaSchulApp}).
\section{Experimental results}
\label{sec:exper}
\begin{figure}[!h]
\centering
\subfigure{
\begin{tikzpicture}[scale=1]
\def\lx{0.1}
\def\ly{1.825}
\begin{axis}[
      width=8.4cm,height=3.4cm,grid=major,grid style = {dotted,black},
      ylabel={\small $\text{$\Delta h_n$}$\;/\;dB $\;\; \rightarrow$},
      ymin=-30, ymax=0,xmin=0,xmax=10,
      ]
      \addplot[thick,blue,solid] table [x index=0, y index=1]{emSPEECH.dat};
      \addplot[thick,green,dashdot] table [x index=0, y index=1]{nlmsSPEECH.dat};
      \addplot[thick,red,dashed] table [x index=0, y index=1]{shSPEECH.dat};
 \end{axis}
  \draw[fill=white] (\lx,\ly) rectangle (6.72+\lx,0.5+\ly);
 \node at (-1.6,2.3) {a)};
 \draw[blue,thick,solid] (0.1+\lx,0.2+\ly) -- +(0.35,0)node[anchor=mid west,black]{\footnotesize EM-NLMS };
 \draw[red,thick,dashed] (2.2+\lx,0.2+\ly) -- +(0.35,0) node[anchor=mid west,black] {\footnotesize Adapt. NLMS};
  \draw[thick,green,dashdot] (4.6+\lx,0.2+\ly) -- +(0.35,0) node[anchor=mid west,black] {\footnotesize Conv. NLMS};
 \end{tikzpicture}}\\[-3.5mm]
 \subfigure{
\begin{tikzpicture}[scale=1]
\def\lx{0.22}
\def\ly{1.52}
\begin{axis}[
      width=8.4cm,height=3.4cm,grid=major,grid style = {dotted,black},
      ylabel={\small $\text{$\alpha_n$}$ $\; \rightarrow$},
      ymin=0, ymax=0.6,xmin=3,xmax=4,
      ]
      \addplot[thick,blue,solid] table [x index=0, y index=1]{emSPEECHstep.dat};
      \addplot[thick,red,dashed] table [x index=0, y index=1]{shSPEECHstep.dat};
            \addplot[thick,green,dashdot] table [x index=0, y index=1]{nlmsSPEECHstep.dat};
 \end{axis}
   \node at (-1.6,1.9) {b)};
 \end{tikzpicture}}\\[-3.5mm]
  \subfigure{
\begin{tikzpicture}[scale=1]
\begin{axis}[
      width=8.4cm,height=3.4cm,grid=major,grid style = {dotted,black},
      ylabel={\small $d_n$ $\; \rightarrow$},
      xlabel={\small time\;/\;s $\;\; \rightarrow$},
      ymin=-1, ymax=1,xmin=3,xmax=4,
      ]
      \addplot[thick,black,solid] table [x index=0, y index=1]{signalSPEECH.dat};
 \end{axis}
   \node at (-1.6,1.9) {c)};
 \end{tikzpicture}}\vspace{-2.3mm}
\caption{Comparison of the EM-NLMS algorithm (``EM-NLMS''), the NLMS algorithm due to \cite{Variable1982} (``Adapt. NLMS``) and the conventional NLMS algorithm (''Conv.~NLMS``) in terms of the system distance~$\Delta h_n$ and the stepsize~\mbox{$\alpha_n$}
(short time frame for visualization purposes) for male speech as input signal (see the microphone signal $d_n$ in Fig.~4c)).}
 \label{fig:ResultsSPEECH}
 \end{figure}
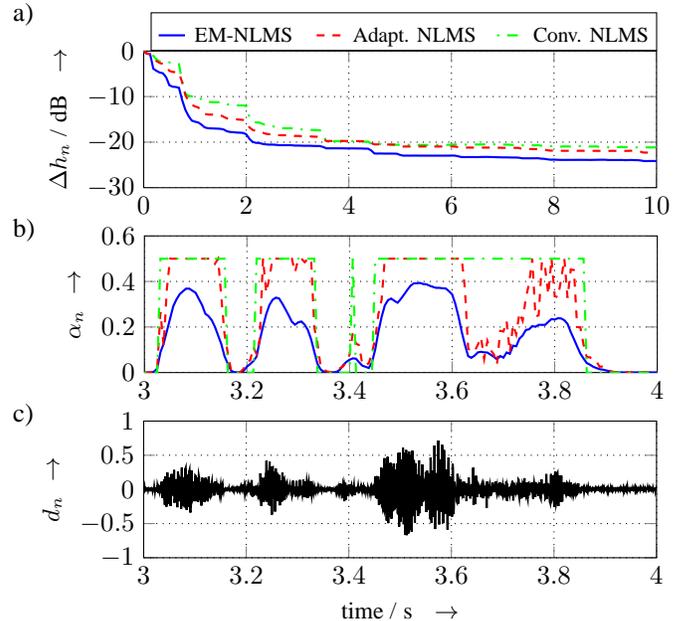

This section focuses on the experimental verification of the EM-NLMS algorithm (``EM-NLMS'') in comparison to the adaptive stepsize-NLMS algorithm described in Section~\ref{sec:Intro} (``Adapt. NLMS'')
and the conventional \ac{NLMS} algorithm (``Conv. NLMS'') with a fixed stepsize.
An overview of the algorithms including the individually tuned model parameters is shown in Table~\ref{tab:Overview}.
Note the regularization of all three stepsize updates by the additive constant~$\epsilon=0.01$ to avoid a division by zero.
For the evaluation, we synthesize the microphone signal by convolution of the loudspeaker signal with an \ac{RIR} vector measured in a room with $T_{60}=100$~ms (filter length $M=512$ at a sampling rate of $16$~kHz). 
This is realized for both white noise and a male speech signal as loudspeaker signals.
Furthermore, background noise is simulated by adding Gaussian white noise at a global signal-to-noise ratio of $20$~dB.
The comparison is realized in terms of the stepsize $\alpha_n$ and the system distance~$\Delta h_n$ as a measure for the system identification performance:\\[-2mm]
\begin{equation}
\Delta h_n= 10 \log_{10}\frac{||\mathbf{\hat{h}}_n-\mathbf{h}_n ||_2^2}{||\mathbf{h}_n ||_2^2} \; \text{dB}, \quad \alpha_n = \lambda_n (\mathbf{x}^T_n \mathbf{x}_n). 
\label{equ:NMA}
\end{equation}\\[-2mm]
The results for white noise as input signal are illustrated in Fig~\ref{fig:ResultsWGN}.
Note that in Fig.~\ref{fig:ResultsWGN}a) the EM-NLMS shows the best system identification compared to the Adapt. NLMS and the Conv.~NLMS.
As depicted in Fig.~\ref{fig:ResultsWGN}b), the stepsize $\alpha_n$ of the EM-NLMS and the Adapt. NLMS decreases from a value of $0.5$ with the stepsize of the EM-NLMS decaying more slowly.\\
For male speech as input signal, we improve the convergence of the Conv. NLMS by setting a fixed threshold to stop adaptation ($\alpha_n=0$) in speech pauses.
Furthermore, the absolute value of $\lambda_n$ for the Adapt. NLMS is limited to 0.5 (for a heuristic justification see~\cite{NLMSschultheis}).
As illustrated in Fig.~\ref{fig:ResultsSPEECH}a), the EM-NLMS shows again the best system identification compared to the Adapt.~NLMS and the Conv.~NLMS. By focusing on a small time frame, we can see in Fig.~\ref{fig:ResultsSPEECH}b)
that the stepsize $\alpha_n$ of the EM-NLMS algorithm
is not restricted to the values of $0$ and $0.5$ (as Conv.~NLMS) and not affected by oscillations (as~Adapt.~NLMS).\\
Note that the only relevant increase in computational complexity of the EM-NLMS relative to the Conv. NLMS is caused by the scalar product $\mathbf{\hat{h}}_n^T\mathbf{\hat{h}}_n$ for the calculation of $C_{\mathbf{w},n}$ (cf. Table~\ref{tab:Overview}), which seems relatively small compared to other sophisticated stepsize adaptation algorithms.
\begin{table}[h!]
\centering
\caption{Realizations of the EM-NLMS algorithm~(``EM-NLMS''), \newline the~NLMS algorithm due to \cite{Variable1982} (``Adapt. NLMS``) and \newline the conventional NLMS algorithm (''Conv.~NLMS``) \qquad \quad}
\vspace{-1mm}
\begin{tabular}[c]{|c|c|}
\hline
 & $\mathbf{\hat{h}}_n = \mathbf{\hat{h}}_{n-1} + \lambda_n  \mathbf{x}_n e_n$   \\
 \hline
EM-NLMS& $\lambda_n= \frac{C_{\mathbf{h},n-1} +C_{\mathbf{w},n}}{\mathbf{x}^T_n\mathbf{x}_n (C_{\mathbf{h},n-1} +C_{\mathbf{w},n}) + C_{v,n} + \epsilon}$   \\[1mm]
& $C_{\mathbf{h},n}= \left( 1 -\lambda_n  \frac{\mathbf{x}^T_n \mathbf{x}_n}{M} \right) (C_{\mathbf{h},n-1} +C_{\mathbf{w},n})$\\[1mm]
& $C_{v,n+1} = \left( d_n - \mathbf{x}^T_n\mathbf{\hat{h}}_n \right)^2 + \mathbf{x}^T_n\mathbf{x}_nC_{\mathbf{h},n}$  \\[1mm]
& $C_{\mathbf{w},n+1} =C_{\mathbf{h},n} - C_{\mathbf{h},n-1} + \frac{\mathbf{\hat{h}}_n^T\mathbf{\hat{h}}_n - \mathbf{\hat{h}}_{n-1}^T\mathbf{\hat{h}}_{n-1}}{M}$ \\[1mm]
& $C_{\mathbf{h},0} =C_{\mathbf{w},0} = C_{v,0} = 0.1$, $\; \epsilon=0.01$ \\[1mm]
\hline
&\\[-6mm]
Adapt. NLMS &$\lambda_{n} \approx  \frac{1}{ N_T}   \frac{  \sum\limits_{\kappa=0}^{N_T-1} \hat{h}^2_{k,n-1} }{ (1-\eta) e_n^2 + \eta \mathcal{E}\{e_{n-1}^2\}+ \epsilon }$\\[1mm]
& $ N_T = 5,\;\; \eta = 0.9,\;\; e_0^2 = 0.1,\;\; \epsilon=0.01   $ \\[1mm]
\hline
Conv. NLMS &$\lambda_{n} =\frac{0.5}{\mathbf{x}^T_n\mathbf{x}_n + \epsilon } , \;\; \epsilon = 0.01$\\[1mm]
\hline
\end{tabular}
\label{tab:Overview}
\end{table}
\section{Conclusion}
\label{sec:conclu}
In this article, we derive the EM-NLMS algorithm from a Bayesian network perspective and show the equality with respect to the NLMS algorithm initially proposed in \cite{Variable1982}.
As main difference, the stepsize is estimated in the M~Step of the EM algorithm instead of being approximated by artificially extending the acoustic echo path.
For the derivation of the EM-NLMS algorithm, which is experimentally shown to be promising for the task of linear AEC, we define a probabilistic model for linear system identification and exploit the product and conditional
independence rules of Bayesian networks.
All together this article exemplifies the benefit of applying machine learning techniques to classical signal processing~tasks.\\[-2.8mm]
\begin{figure}[!h]
\centering
\subfigure{
\begin{tikzpicture}[scale=1]
\def\lx{0.1}
\def\ly{1.825}
\begin{axis}[
      width=8.4cm,height=3.4cm,grid=major,grid style = {dotted,black},
      ylabel={\small $\text{$\Delta h_n$}$\;/\;dB $\;\; \rightarrow$},
      ymin=-50, ymax=0,xmin=0,xmax=5,
      ]
      \addplot[thick,blue,solid] table [x index=0, y index=1]{emWGN.dat};
      \addplot[thick,green,dashdot] table [x index=0, y index=1]{nlmsWGN.dat};
      \addplot[thick,red,dashed] table [x index=0, y index=1]{shWGN.dat};
 \end{axis}
 \draw[fill=white] (\lx,\ly) rectangle (6.72+\lx,0.5+\ly);
 \node at (-1.6,2.3) {a)};
 \draw[blue,thick,solid] (0.1+\lx,0.2+\ly) -- +(0.35,0)node[anchor=mid west,black]{\footnotesize EM-NLMS };
 \draw[red,thick,dashed] (2.2+\lx,0.2+\ly) -- +(0.35,0) node[anchor=mid west,black] {\footnotesize Adapt. NLMS};
  \draw[thick,green,dashdot] (4.6+\lx,0.2+\ly) -- +(0.35,0) node[anchor=mid west,black] {\footnotesize Conv. NLMS};
 \end{tikzpicture}}\\[-2mm]
 \subfigure{
\begin{tikzpicture}[scale=1]
\def\lx{0.1}
\def\ly{1.825}
\begin{axis}[
      width=8.4cm,height=3.4cm,grid=major,grid style = {dotted,black},
      ylabel={\small $\text{$\alpha_n$}$ $\; \rightarrow$},
      xlabel={\small time\;/\;s $\;\; \rightarrow$},
      ymin=0, ymax=0.6,xmin=0,xmax=5,
      ]
      \addplot[thick,blue,solid] table [x index=0, y index=1]{emWGNstep.dat};
      \addplot[thick,red,dashed] table [x index=0, y index=1]{shWGNstep.dat};
      \addplot[thick,green,dashdot] table [x index=0, y index=1]{nlmsWGNstep.dat};
 \end{axis}
  \node at (-1.6,1.9) {b)};
 \end{tikzpicture}}\vspace{-2.3mm}
\caption{Comparison of the EM-NLMS algorithm (``EM-NLMS''), the NLMS algorithm due to \cite{Variable1982} (``Adapt. NLMS``) and the conventional NLMS algorithm (''Conv.~NLMS``) in terms of the system distance~$\Delta h_n$ and the stepsize \mbox{$\alpha_n$}
for white Gaussian noise~as~input~signal.}
 \label{fig:ResultsWGN}
 \end{figure}
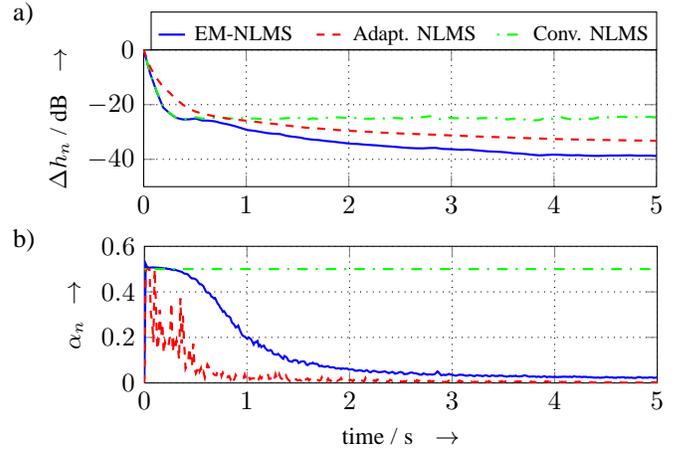\\[-7mm]
  \newpage

\bibliographystyle{IEEEbib}
\bibliography{literature}

\end{document}